\documentclass{article}

\usepackage{times}
\usepackage{algorithm} 
\usepackage{algorithmic} 
\usepackage{booktabs}
\usepackage{multirow} 
\usepackage{amsmath}
\usepackage{graphicx}
\usepackage{url}
\usepackage{color}

\title{An Active Learning Approach for Jointly Estimating Worker Performance and Annotation Reliability with Crowdsourced Data}

\author{
\begin{tabular}{ccc}
Liyue Zhao$^1$ & Yu Zhang$^2$ & Gita Sukthankar$^1$\\
lyzhao@cs.ucf.edu & yuz1988@iastate.edu & gitars@eecs.ucf.edu
\end{tabular}
\\
$^1$ Department of EECS, University of Central Florida\\ 
$^2$ Department of Computer Science, Iowa State University
}
\begin{document}

\maketitle

\begin{abstract}
Crowdsourcing platforms offer a practical solution to the problem of affordably annotating large datasets for training supervised classifiers.  Unfortunately, poor worker performance frequently threatens to compromise annotation reliability, and requesting multiple labels for every instance can lead to large cost increases without guaranteeing good results.   Minimizing the required training samples using an active learning selection procedure reduces the labeling requirement but can jeopardize classifier training by focusing on erroneous annotations.   This paper presents an active learning approach in which worker performance, task difficulty, and annotation reliability are jointly estimated and used to compute the risk function  guiding the sample selection procedure.  We demonstrate that the proposed approach, which employs active learning with Bayesian networks, significantly improves training accuracy and correctly ranks the expertise of unknown labelers in the presence of annotation noise. 
\end{abstract}

\section{Introduction}

Our work is motivated by the recent interest in the use of crowdsourcing~\cite{howe2006rise} as a source of annotations from which to train machine learning systems. Crowdsourcing transforms the problem of generating a large corpus of labeled real-world data from a monolithic labor-intensive ordeal into a manageable set of small tasks, processed by thousands of human workers in a timely and affordable manner.  Services such as Amazon's Mechanical Turk (MTurk) have made it possible for researchers to acquire sufficient quantities of labels, enabling the development of a variety of applications driven by supervised learning models. However, employing crowdsourcing to label large quantities of data remains challenging for two important reasons: limited annotation budget and label noise.  First, although the unit cost of obtaining each annotation is low, the overall cost grows quickly since it is proportional to the number of requested labels, which can number in the millions.  This has stimulated the use of approaches such as active learning~\cite{settles2010active} that aim to minimize the amount of data required to learn a high-quality classifier. Second, the quality of crowdsourced annotations has been found to be poor~\cite{sheng2008get}, with causes ranging from workers who overstate their qualifications, lack of motivation among labelers, haste and deliberate vandalism.  Unfortunately, the majority of popular active learning algorithms, while robust to noise in the input features, can be very sensitive to label noise, necessitating the development of approaches specifically designed for noisy annotations.  In this paper, we focus on the causal factors behind noisy crowdsourced annotations, worker quality and task difficulty.

A common simplifying assumption is that of \emph{identicality}: all labelers provide annotations with the same accuracy and all annotation tasks pose the same difficulty to all workers.  Under this assumption, a good way to combat annotation noise is to request multiple labels for each selected task and then to apply majority voting.  The simplest approach of requesting the same number of labels for each instance is not usually the most cost-effective since label redundancy increases the overall cost by a multiplicative factor of at least three. 

Better results can be obtained by applying weighted voting which assigns different weights to labelers based on their previous performance~\cite{sheng2008get,zhao2011incremental,zheng2010active}. In this paper, we use the Generative model of Labels, Abilities, and Difficulties~\cite{whitehill2009whose} in which labeler expertise and the task difficulty are simultaneously estimated using EM (Expectation-Maximization) to learn the parameters of a probabilistic graphical model which represents the relationship between labelers, tasks and annotation predictions.   Rather than using previous performance to assign weights, the estimated labeler expertise is used to allocate weights to annotator votes.   This paper focuses on the problem of reducing the label budget used by the GLAD model with active learning.

Theoretically the most straightforward and effective strategy for active learning is to select samples that offer the greatest reductions to the risk function. Thus, ``aggressive'' criteria such as least confidence, smallest margin, or maximum entropy can enable active learning to obtain high accuracy using a relatively small number of labels to set the decision boundary.  Unfortunately, the existence of label noise can trigger failures in aggressive active learning methods because even a single incorrect label can cause the algorithm to eliminate the wrong set of hypotheses, thus focusing the search on a poor region of the version space.  In contrast, the proposed combination of the probabilistic graphical model and active learning, avoids explicitly eliminating any set of hypotheses inconsistent with the label provided by annotators.

Specifically, this paper makes two contributions:
\begin{enumerate}
\item we propose a new sampling strategy which iteratively selects the \emph{combination} of worker and task which offers the greatest risk reduction between the current labeling risk and the expected posterior risk.  The strategy aims to focus on sampling reliable labelers and uncertain tasks to train the Bayesian network.
\item we present comprehensive evaluations on both simulation and real world datasets that show not only the strength of our proposed approach in significantly reducing the quantity of labels required for training the model, but also the scalability of our approach at solving practical crowdsourcing tasks which suffer large amounts of annotation noise.
\end{enumerate}

\section{Related Work}

Howe et al. ~\cite{howe2006rise} coined the phrase ``crowdsourcing'' to describe the concept of outsourcing work to a cheap labor pool composed of everyday people who use their spare time to create content and solve problems. Doan et al. ~\cite{doan2011crowdsourcing} define crowdsourcing as enlisting a crowd of humans to help solve a problem defined by the system owners. Crowdsourcing annotation services, such as Amazon's Mechanical Turk, have become an effective way to distribute annotation tasks over multiple workers~\cite{vijayanarasimhan2010far}; however, Sheng et al.~\cite{sheng2008get} noted the problem that crowdsourcing annotation tasks may generate unreliable labels. 

Several works~\cite{donmez2010probabilistic,yan2011active,raykar2009supervised} propose different approaches to model the annotation accuracy of workers; all of these approaches assume there are multiple experts/annotators providing labels but that no oracle exists. Donmez et al.~\cite{donmez2010probabilistic} propose a framework to learn the expected accuracy at each time step. The estimated expected accuracies are then used to decide which annotators should be queried for a label at the next time step. Yan et al.~\cite{yan2011active} focus on the multiple annotator scenario where multiple labelers with varying expertise are available for querying. This method can simultaneously answer questions such as which data sample should be labeled next and which annotator should be queried to benefit the learning model. Raykar et al.~\cite{raykar2009supervised} use a probabilistic model both to evaluate the different experts and also to provide an estimate of the actual hidden labels.

Our work is strongly influenced by GLAD~\cite{whitehill2009whose}, which uses a probabilistic model to simultaneously estimate the labels, the labeler expertise, and the task difficulty which are represented as latent variables in the Bayesian network. The model can estimate the label of a new task with a weighted combination of labels from different labelers based on their expertise inferred in the training phase.  However, the original GLAD model does not use active learning, unlike our proposed approach which offers substantial reductions to the labeling cost, without sacrificing annotation accuracy.

Tong and Koller\cite{tong2001active_b} proposed a way to implement active learning in Bayesian networks with a simple structure. Tong and Koller chose Kullback-Leibler divergence as the loss function to measure the distance between distributions. Their algorithm iteratively computes the expected change in risk and makes the sample query with the greatest expected change. This strategy is guaranteed to request the label of the sample that reduces the expected risk the most, but does not account for worker performance.  In this paper, we propose an alternate selection strategy which selects pairs of workers and samples and uses an entropy-based risk function.

\section{Method}

In this section, we describe our active learning approach for jointly estimating worker performance and annotation reliability.  The first part of the section defines the probabilistic graphical model for estimating the expertise of labelers and the difficulty of annotation tasks before describing how EM is used to estimate the model parameters.  The second subsection introduces our active learning approach for sampling workers and tasks using an entropy-based risk function.

\subsection{Probabilistic Graphical Model}


Our work utilizes the generative model proposed by Whitehill et al.~{{whitehill2009whose}. The structure of the graphical model is shown in Figure~\ref{fig:bn}. The same model is used to estimate the expertise of workers, the difficulty of annotation tasks, and the true labels. The expertise of worker $i$ is defined as $\alpha_i \in (-\infty,+\infty)$, which corresponds to the worker's level of annotation accuracy.  As $\alpha_i$ approaches $+\infty$, worker $i$ becomes an increasingly capable oracle who almost always gives the correct label and as $\alpha_i$ approaches $-\infty$ the worker almost always provides the wrong label. $\alpha_i=0$ means the worker has no capability to distinguish the difference between the two classes and just randomly guesses a label. The difficulty of task $j$ is parameterized by $\beta_j$ where $\frac{1}{\beta_j} \in (0,+\infty)$.  For easy tasks, $\frac{1}{\beta_j}$ approaches zero, and it is assumed that practically every worker can give the correct label for the task.  As $\frac{1}{\beta_j}$ approaches $+\infty$, the task becomes so difficult that almost no one is able to provide the right answer. In this paper, we only consider binary classification problems which assume that both the true label $Z_j$ and the annotation $l_{ij}$ provided by the worker $i$ are binary labels. $l_{ij} \in \{-1,+1\}$ is defined as the label of task $j$ provided by annotator $i$.  This is the only observable variable in the graphical model. Since in most crowdsourcing platforms, it is unlikely that the same labelers will be responsible for annotating all tasks in the dataset, this model also works well when the observation is incomplete. The true label $z_j$ of task $j$ is the variable that we are going to estimate to evaluate the labeling performance of the model. 

\begin{figure}[t]
\includegraphics[width=1\columnwidth]{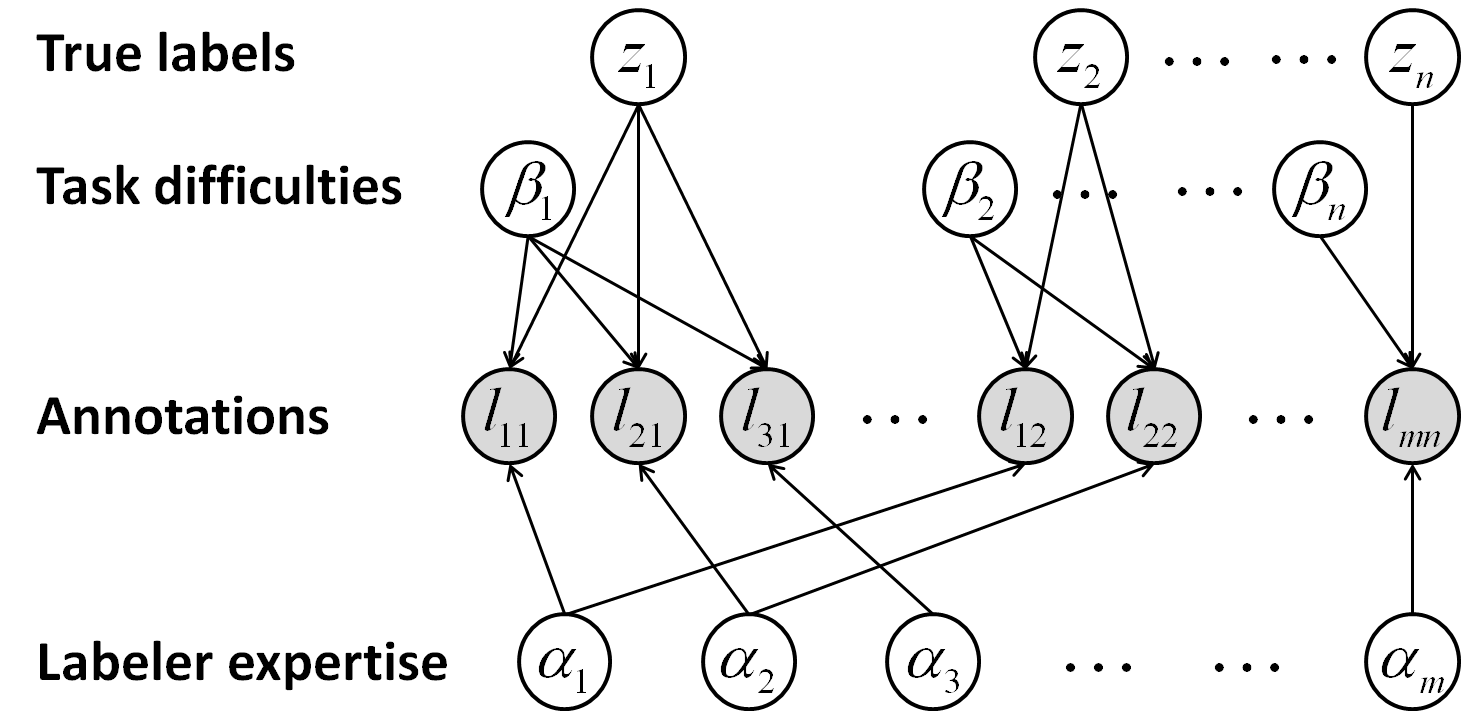}
\caption{The structure of the Bayesian networks used to estimate the true labels, worker expertise, and annotation task difficulty}
\label{fig:bn}
\end{figure}

Given the definitions above, the probability of annotator $i$ providing the correct label for task $j$ is defined in Equation~\ref{equ:posti}. For a more skilled labeler with a larger $\alpha$, or an easier task with a larger $\beta$, the probability of providing the correct label should be larger. However, if the task is too difficult ($\frac{1}{\beta} \rightarrow +\infty$) or the labeler has no background in performing the task ($\alpha = 0$), the labeler can only give a random guess ($p=0.5$) about the task label.

\begin{equation}
P(l_{ij} = z_{j}|\alpha_i,\beta_j) = \frac{1}{1+e^{-\alpha_i \beta_j}}
\label{equ:posti}
\end{equation}

The set of parameters of the graphical model $\pmb{\alpha} = \{\alpha_1,...,\alpha_i,...,\alpha_m\}$ and $\pmb{\beta} = \{\beta_1,...,\beta_j,...,\beta_n\}$ are represented as $\pmb{\theta} = \{\pmb{\alpha}, \pmb{\beta}\} $. $\mathbf{L}$ and $\mathbf{z}$ are defined as the set of labels provided by the workers (observed) and the true labels (not known).

EM is an iterative algorithm for maximum likelihood estimation in a graphical model with incomplete data.  In this annotation task, the learning procedure starts with the set of unreliable annotations $\mathbf{L}$. The EM algorithm iteratively estimates the unknown parameters $\pmb{\theta}$ with the current observations $\mathbf{L}$ and then updates the belief of the true labels $\mathbf{z}$ using these estimated parameters.  The description of the EM algorithm as applied to this model is given below.

\begin{enumerate}
	\item[E-Step:] At the expectation step, the parameters $\pmb{\theta}$ estimated in the maximization step are fixed. The posterior probability of $z_j \in \{0,1\}$ given $\pmb{\theta}$ is computed as:
	\begin{equation}
	p(z_i|\mathbf{L},\pmb{\theta}) \propto p(z_j) \prod_i p(l_{ij},\alpha_i,\beta_j)
	\label{equ:estep}
	\end{equation}
	\item[M-Step:] Given the posterior probability of $z_j$, the maximization step uses the cost function $Q(\pmb{\theta})$ to estimate a locally optimal solution of parameters $\pmb{\theta}$:
	\begin{equation}
	Q(\pmb{\theta}) = \sum_j E[\ln{p(z_j)}] + \sum_{ij} E[\ln{p(l_{ij}|z_j,\alpha_i,\beta_j)}]
	\label{equ:mstep}
	\end{equation}
\end{enumerate}

The aim of the EM algorithm is to return the parameters $\pmb{\theta^{*}}$ which maximize the function $Q(\pmb{\theta})$:

\begin{equation}
\pmb{\theta}^{*} = \arg \max_{\pmb{\theta}} Q(\pmb{\theta})
\label{equ:theta}
\end{equation}

$\pmb{\theta^{*}}$ represents the optimal estimate of the expertise of labelers and the difficulty of tasks with the current set of annotations.



\subsection{Query-by-Uncertainty}
The essential question in active learning is which criterion performs the best at identifying the most valuable unlabeled sample under the present parameters.  In this paper, we employ the entropy of the label distribution to represent the risk of a particular label assignment. We assume the class of labels is defined as $\mathcal{C} = \{-1,+1\}$, and the risk function is

\begin{equation}
Risk(z_j) = -\sum_{|\mathcal{C}|} \frac{p(z_j|\mathbf{L},\pmb{\theta^{*}})}{|\mathcal{C}|} \log{ \frac{p(z_j|\mathbf{L},\pmb{\theta^{*}})}{|\mathcal{C}|} }
\label{equ:risk}
\end{equation}

\noindent where $z_j$ is the true label and $p(z_j|\mathbf{L},\pmb{\theta^{*}})$ represents the probability of annotating task $j$ with label $z_j$ under the present parameters $\pmb{\theta^{*}}$. This risk function evaluates the risk of assigning sample $j$ label $z_j$. Our active learning algorithm preferentially selects the sample that maximizes this risk function to be annotated.

The algorithm for using active learning within the Bayesian network is shown in Algorithm~\ref{alg:albn}. The algorithm begins with the data to be annotated and a partially completed label matrix $\mathbf{L}$. The set of parameters $\pmb{\theta}$ is initialized at this stage, and the labeling budget for the crowdsourcing is allocated.  The goal of this algorithm is to learn the set of optimal parameters $\pmb{\theta^*}$ with a given label budget.  At each iteration, our algorithm selects the sample that maximizes the risk function defined in Equation~\ref{equ:risk} with the parameters $\pmb{\theta^*}$ estimated using the EM algorithm. The sample is then annotated by requesting label $l_{ij}$ from the labeler $i$ with the maximum current $\alpha_i$ who had not previously contributed a label to that task.

\begin{algorithm}[t]
\caption{Active Learning Algorithm}
\label{alg:albn} 
\small
\begin{algorithmic}[1]
\REQUIRE ~~\ \\
\textbf{Input:}\\
A set of data along with a matrix of partial labels $\mathbf{L}$\\
The initial set of parameters $\pmb{\theta}$\\
$B$: labeling budget.\\
\textbf{Output:}\\
$\pmb{\theta}^* = \{ \pmb{\alpha}^*, \pmb{\beta}^* \}$: The set of parameters representing the expertise of labelers and the difficulty of tasks.\\
\WHILE {$B > 0$}
  \STATE Use the EM algorithm to update the parameters $\pmb{\theta} = \{ \pmb{\alpha}, \pmb{\beta} \}$ using Equations~\ref{equ:estep} and \ref{equ:mstep};\\
  \STATE Find parameters $\pmb{\theta}^{*}$ that maximize the function $Q(\pmb{\theta})$ using Equation~\ref{equ:theta};\\
  \STATE Calculate the risk of assigning label $z_j$ to task $j$ with Equation~\ref{equ:risk};\\
  \STATE Query the label $l_{ij}$ by assigning task $j$ with the maximum risk to the worker $i$ with the highest $\alpha_i$ and $l_{ij}\ne 0$;\\
  \STATE $B \leftarrow B-1$;\\
\ENDWHILE \\
\STATE Return the optimal parameters $\pmb{\theta}^*$.
\end{algorithmic}
\end{algorithm}

\section{Worker Selection}

Selecting the task to be labeled could be easily solved by applying active learning strategies to identify the most uncertain or informative task to workers. However, how to select the right worker to annotate the task has became another issue which is out of the scope of active learning. Selecting the most ``uncertain" worker may polish the estimation of the worker expertise, but simultaneously reduce the speed to improve the training accuracy, since such strategy inevitably pick workers who the system is most unfamiliar with, rather than the best workers. 

The straightforward way we have used in Algorithm~\ref{alg:albn} is to ask the worker $i$ with the highest expertise estimation $\alpha^*_i$ to provide the label. The experiment results shows this strategy works well. However, there exists some arguments that challenge the strategy since 1) $\alpha^*_i$ is only the estimation of $\alpha_i$, which means picking the worker with largest  $\alpha^*$ may ignore the real ``best worker" and 2) the evaluation of other workers are ignored since they will not be selected forever. The goal of this section is to investigate the performances of different worker selection strategies.

Beside always sampling the best worker, we evaluate two other options: weighted selection and $\epsilon$-greedy selection. For the weighted selection strategy, the probability of selecting worker $i$ is proportional to the expertise $\alpha^*_i$. 

\begin{equation}
p(i) = \frac{\alpha^*_i}{\sum_i{\alpha^*_i}}
\label{equ:wghs}
\end{equation}

An alternative worker selection strategy is the $\epsilon$-greedy selection algorithm, which has been used in studying the exploration-exploitation tradeoff in reinforcement learning~\cite{kuleshov2010algorithms}. This algorithm selects the most possible worker $i$ (with the highest expertise $\alpha_i$) with probability $1 - \epsilon$. and selects other workers with probability $\epsilon$.

\begin{equation}
p(i) =\left\{
\begin{array}{ll}
1 - \epsilon + \frac{ \epsilon }{m}  & i = \max{\alpha^*_i} \\
\frac {\epsilon}{m}   & otherwise \\
\end{array}
\right.
\label{equ:epsilon}
\end{equation}

\noindent where $\epsilon \in [0,1)$ represents how many weights to take from selecting the best worker to other workers, and $m$ is the number of workers available to the task in total.

\section{Experiments}

The aim of our experiments is to demonstrate that, in cases where the tasks are not labeled by all labelers, our proposed sampling strategy in choosing samples to be labeled compared with the random sampling strategy. Although there are three sets of parameters in the model: $\alpha_i$ which is the expertise of the labeler $i$, $\beta_j$ which is the difficulty of the task $j$ and $z_j$ which is the true label of the task $j$, our proposed method proves that focusing on estimating a better labeler expertise $\pmb{\alpha}$ is more important than the task difficulty $\pmb{\beta}$ in predicting the real label of the task $\mathbf{z}$. Compared with random sampling, given the same number of training labels, active sampling has advantages in 1) predicting more correct labels and 2) identifying a more correlated rank of labelers.

In the following experiments, we test the performance of our proposed active learning algorithm on 1) a pool of simulated workers and tasks and 2) a standard image dataset crowdsourced by workers on Mechanical Turk.  The algorithm is evaluated using the annotation accuracy of the predicted labels $\mathbf{z}$ and by comparing the actual and predicted labeler expertise $\pmb{\alpha}$ with $\pmb{\alpha}'$ using both Spearman rank and Pearson correlation statistics.   We omit results for evaluating the model's predictions of task difficulty, since its main importance is its contribution to label noise which we are measuring directly through annotation accuracy.

Our experiments use the same evaluation protocol and dataset as Whitehill et al.~\cite{whitehill2009whose} while attempting to reduce the labeling budget required to reach the desired accuracy level using three different active learning strategies:
\begin{description}
\item[proposed:] selects the most capable worker and most uncertain task using Algorithm~\ref{alg:albn};
\item[traversal:] sequentially selects tasks and randomly selects workers;
\item[random:] randomly selects pairs of workers and tasks.
\end{description}

To investigate the performance of worker selection, we implement our proposed active learning approach with different worker selection strategies on both simulated workers and real workers on Mechanical Turk.  The experiments compare worker expertise using both Spearman rank and Pearson correlation for simulated workers and the annotation accuracy to evaluate real workers. Our experiments attempt to identify good workers by using four different worker selection strategies:

\begin{description}
\item[best worker:] the worker $i$ with highest parameter $\alpha_i$ will be selected;
\item[weighted selection:] the probability of the worker $i$ to be selected is proportional to the value of $\alpha_i$;
\item[epsilon=0.5:]  the $\epsilon$-greedy selection algorithm (Equation~\ref{equ:epsilon}), with $\epsilon=0.5$;
\item[epsilon=0.1:]  the $\epsilon$-greedy selection algorithm (Equation~\ref{equ:epsilon}), with $\epsilon=0.1$.
\end{description}

At the initialization phase of the active learning, all tasks are annotated by exactly two labelers who are randomly selected for each task to seed the training pool.  Using this pool, $\pmb{\alpha}$ and $\pmb{\beta}$ are estimated. At each iteration, one sample is selected from the pool of unlabeled pairs (labeler and task). The selected sample is then added to the training pool, and the parameters $\pmb{\alpha}$, $\pmb{\beta}$ and $\pmb{z'}$ are updated using the extended training pool.

\subsubsection{Simulated Worker Pool}

Since it is difficult to definitively determine the skill level of real human labelers, we ran an initial set of experiments using a simulated worker pool.  In the first set of experiments, we evaluate a simple population that consists of workers with only two skill levels and task with two difficulty levels.  The pool of simulated workers consists of 24 labelers (8 good, 16 bad) annotating 750 tasks (500 easy, 250 hard).  The probabilities for each type of labeler correctly annotating different types of tasks are shown in Table~\ref{tab:binary}.  

\begin{table}
\caption{Annotation accuracy of simulated workers}
\label{tab:binary}
\begin{center}
\begin{tabular}{lll}
\toprule
Type of Labeler	& Hard Task	& Easy Task\\
\midrule
\hline
Good	& 0.95	& 1 \\
Bad	& 0.54	& 1 \\
\bottomrule
\end{tabular}
\end{center}
\end{table}

At the initialization stage, each task is labeled by $2$ randomly selected labelers, which produces a pool of 1500 labels before all sample selection strategies start running.  Each strategy runs for 4000 iterations.  Figure~\ref{fig:exp1acc} shows a comparison of the training accuracy of the different sampling strategies (proposed, traversal, and random). Since the binary case is relatively simple, the training accuracy starts from a high level ($80\%$) with the 1500 initial samples. Our proposed strategy improves rapidly and converges to almost $100\%$ training accuracy in about 3300 $(18.3\%)$ labels. The traversal strategy barely converges to the same accuracy level with 5500 $(30.6\%)$ labels.  Unsurprisingly the random strategy performs the worst and after selecting 5500 labels, it reaches a $94\%$ training accuracy.

\begin{figure}[t]
\includegraphics[width=1\columnwidth]{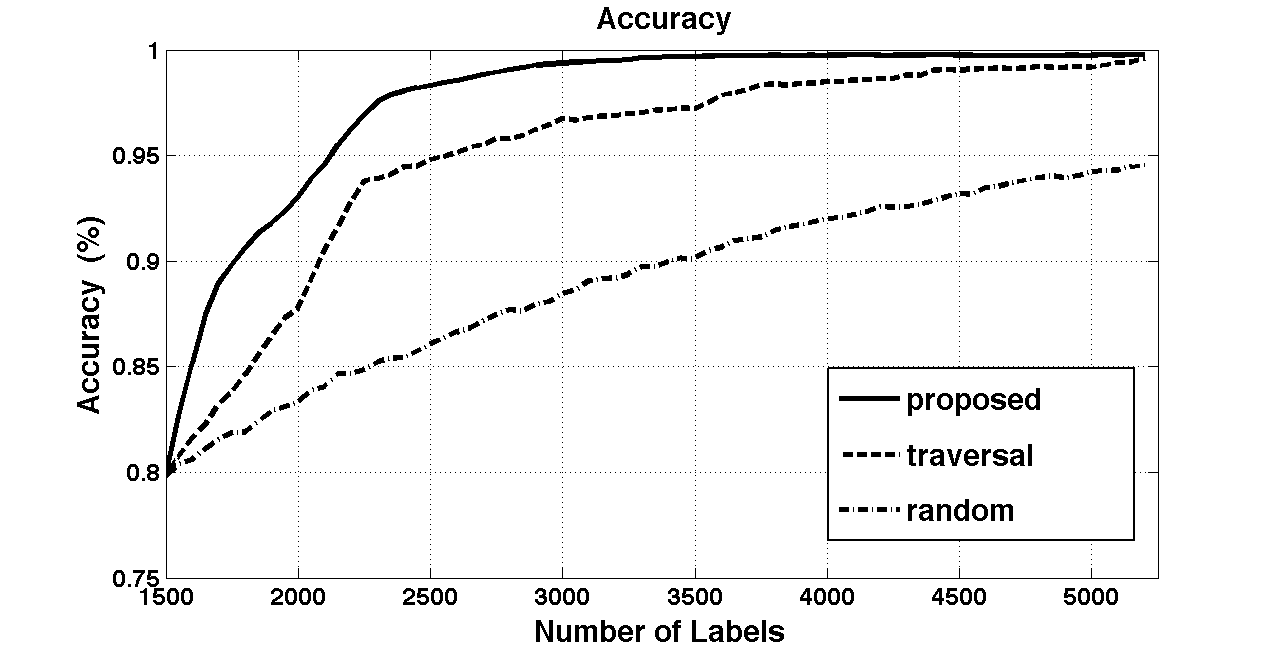}
\caption{Comparison of the annotation accuracy of different sampling strategies (proposed, traversal, random) using the simulated worker pool with binary worker performance levels.}
\label{fig:exp1acc}
\end{figure}

\begin{figure}[t]
\includegraphics[width=\columnwidth]{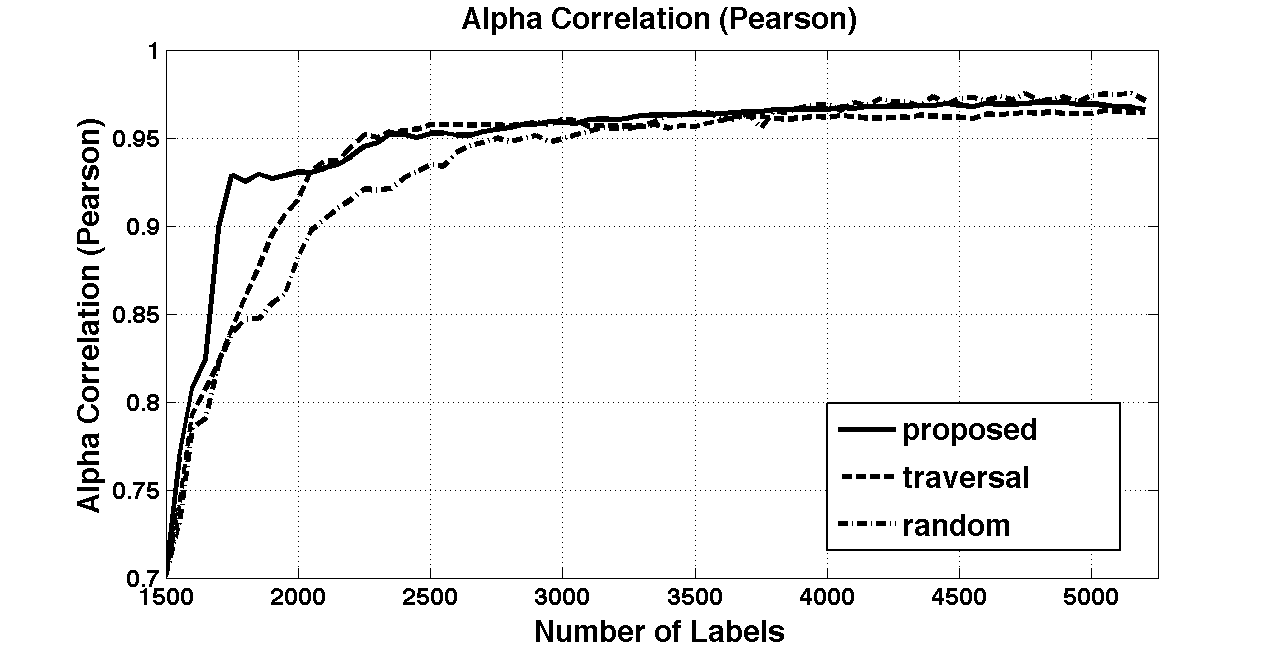}
\caption{Comparison of the different sampling strategies at estimating worker performance in the simulated pool with binary worker performance levels using Pearson correlation.}
\label{fig:exp1pea}
\end{figure}

\begin{figure}[t]
\includegraphics[width=1\columnwidth]{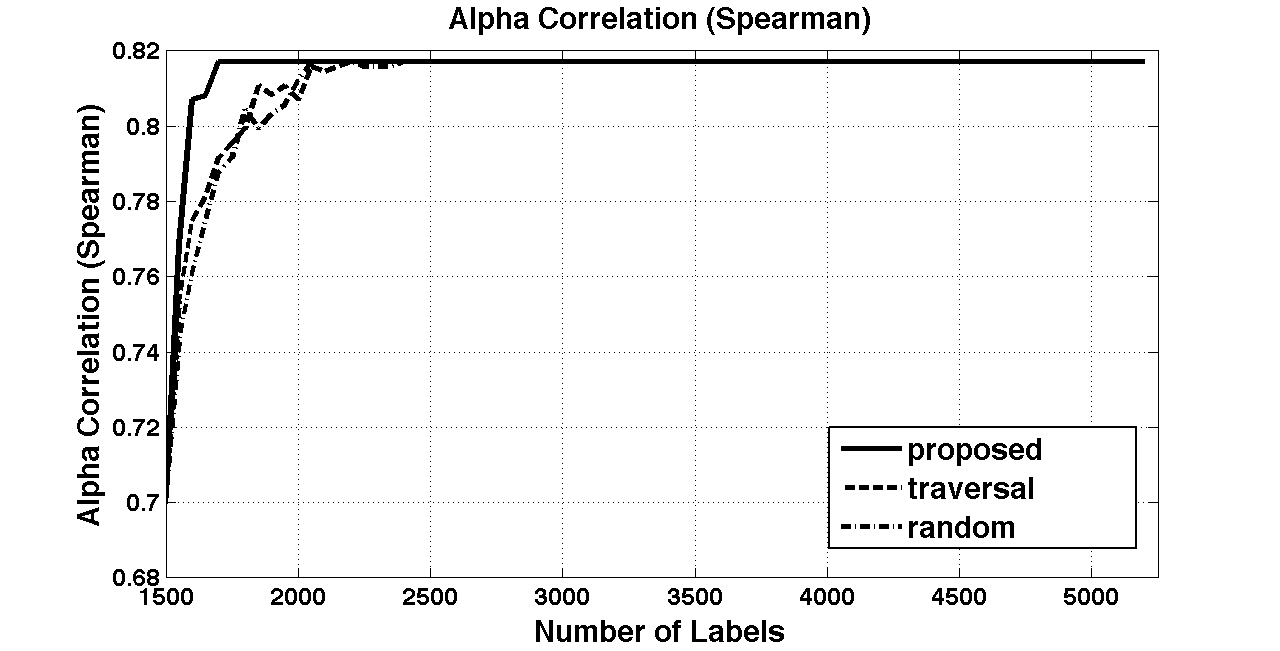}
\caption{Comparison of different sampling strategies at estimated worker performance in the simulated pool with binary worker performance levels using Spearman correlation.}
\label{fig:exp1spe}
\end{figure}

Figure~\ref{fig:exp1pea} and Figure~\ref{fig:exp1spe} shows the result of using Pearson and Spearman rank correlations to evaluate the estimation of $\alpha$. Both Pearson correlation and Spearman rank correlation measure the strength of a relationship between two sets of variables. A perfect correlation of 1 represents two set of variables that are perfectly correlated with each other.   Pearson correlation measures the correlation between the actual estimates of worker performance, whereas the Spearman rank correlation compares the similarity of the relative rankings.   For practical purposes, having a good Spearman rank correlation is sufficient for correctly selecting the best labeler from the pool.

 The goal of this experiment is to evaluate the estimate of labeler expertise $\alpha$ compared to the real value $\alpha'$. The result shows, by using our proposed active learning strategy, the correlation rapidly jumps to a high score only $200$ iterations after initialization. Although the random and traversal sampling strategies reach the same level of correlation in the Pearson correlation, the active learning strategy wins a overwhelming victory with the Spearman rank correlation. The correlation score jumps to the optimal value after only $200$ labels, compared to random and traversal who query more than $500$ labels to approach the comparable correlation score. (Notice that the optimal value of the Spearman rank correlation is as low as 0.82 because there are many ties in the real rankings which inflate the differences between the real ranking and the estimated ranking.)

Figure~\ref{fig:exp1peaworker} and Figure~\ref{fig:exp1speworker} show the result of using Pearson and Spearman rank correlations to evaluate the performance of different worker selection strategies. The results show that no worker selection strategies shows an overwhelming advantage in this simple binary classification case.

\begin{figure}[t]
\includegraphics[width=\columnwidth]{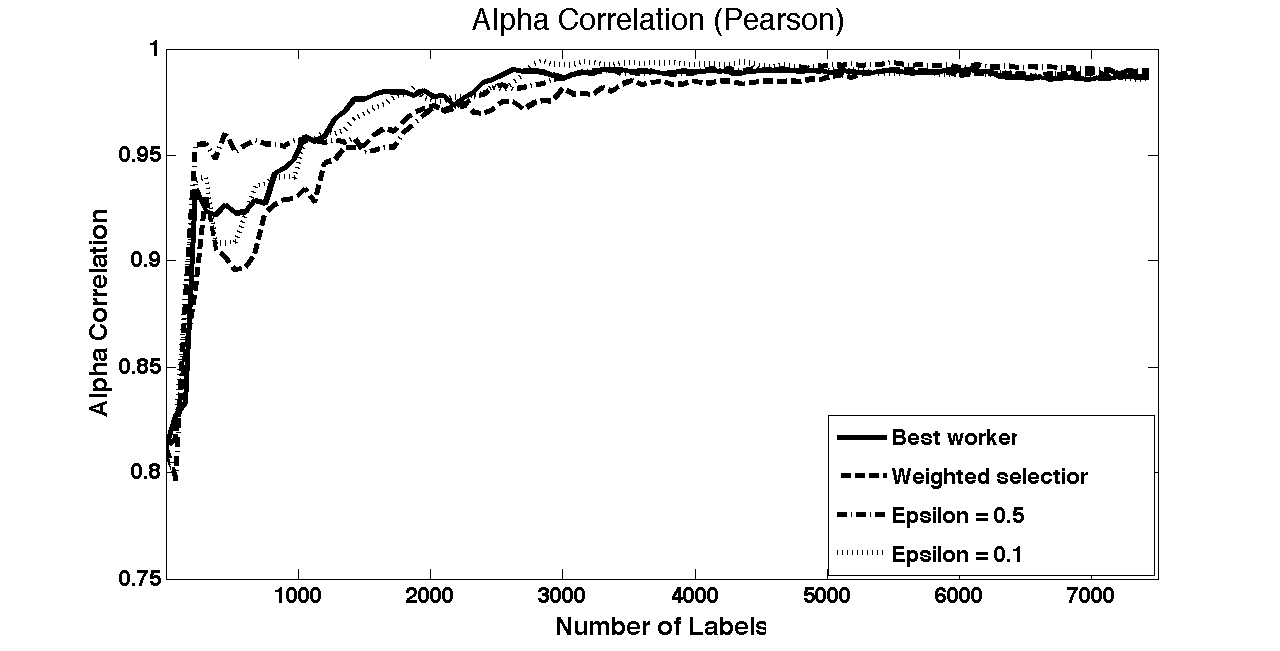}
\caption{Comparison of the different worker selection strategies at estimating worker performance in the simulated pool with binary worker performance levels using Pearson correlation.}
\label{fig:exp1peaworker}
\end{figure}

\begin{figure}[t]
\includegraphics[width=1\columnwidth]{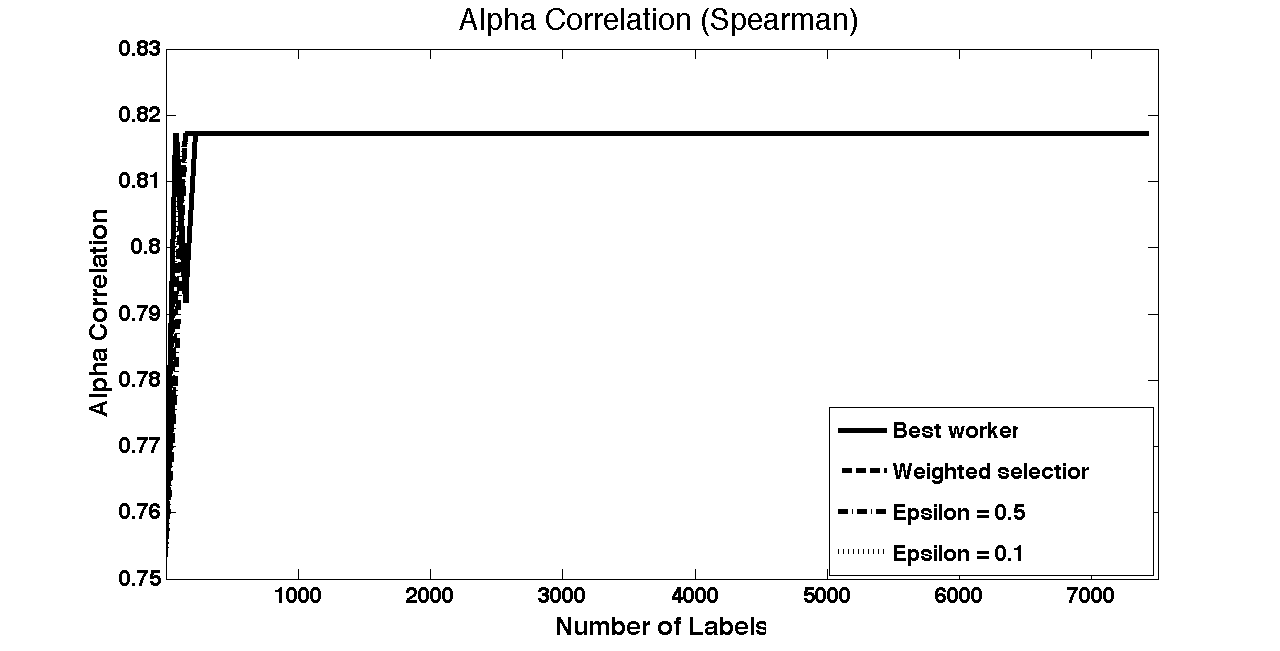}
\caption{Comparison of different worker selection strategies at estimated worker performance in the simulated pool with binary worker performance levels using Spearman correlation.}
\label{fig:exp1speworker}
\end{figure}

However, in the real world there can be many different levels of labeler expertise and task difficulty which makes the estimation problem more challenging.  To model this, in the second experiment, the worker pool is simulated using $\alpha'$ and $\beta'$ that are generated from a Gaussian distribution to simulate a more diverse population of labelers and tasks. We create $50$ labelers and $1000$ tasks in this experiment.  The expertise of labeler $i$ is determined by $\alpha'_i \sim \mathcal{N}(1,1)$, and the difficulty of task $j$ is determined by $\beta'_j \sim \mathcal{N}(1,1)$.  At the initialization stage, each task was labeled by two randomly selected labelers, which yields 2000 labels before all strategies start running. Each strategy runs for 10000 iterations.

Figure~\ref{fig:exp2acc} shows the training accuracy of different sampling strategies (proposed, traversal, and random).  Our proposed strategy still performs strongly at estimating the training accuracy which converges to $97\%$ with 10000 labels. The traversal strategy not only converges slower but reaches a slightly lower accuracy rate around $96\%$ with 10000 labels. The random strategy still performs the worst. After selecting 12000 labels, it reaches a $92\%$ training accuracy.

\begin{figure}[!htb]
\includegraphics[width=1\columnwidth]{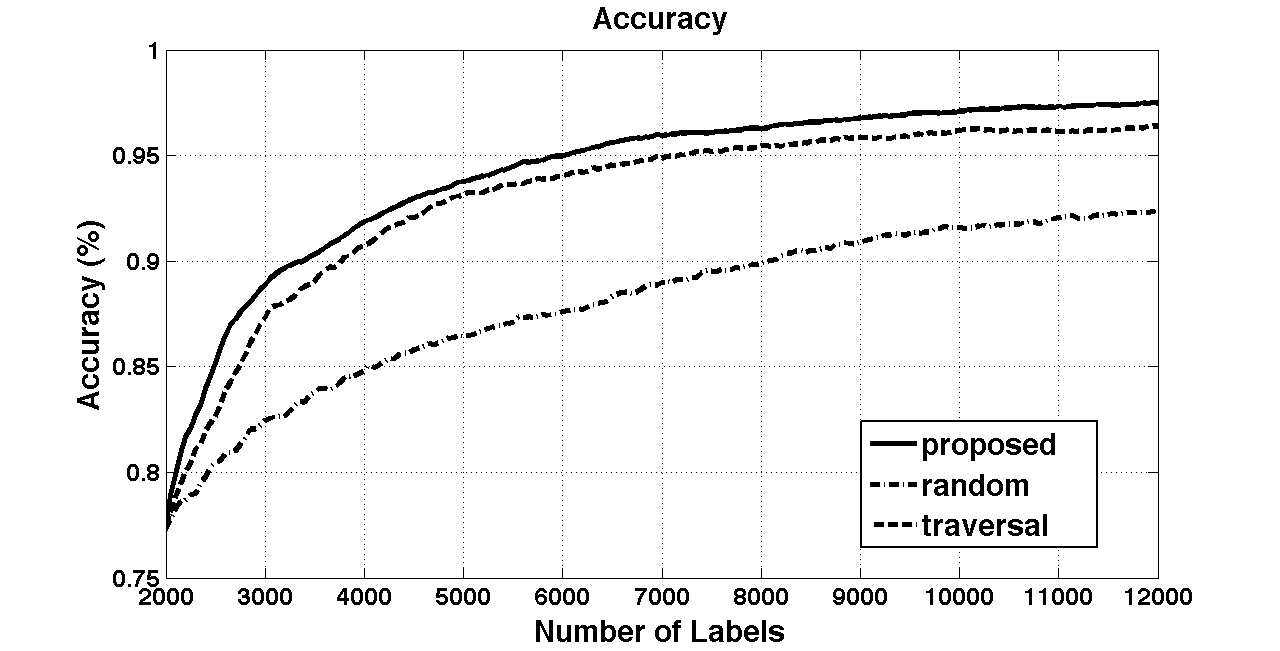}
\caption{Comparison of the annotation accuracy of different sampling strategies (proposed, traversal, random) using the simulated worker pool with Gaussian worker performance levels.}
\label{fig:exp2acc}
\end{figure}

Figure~\ref{fig:exp2pea} and Figure~\ref{fig:exp2spe} show the result of using Pearson and Spearman rank correlation to evaluate the estimate of $\alpha$. The results show that our active learning doesn't show an overwhelming advantage as in the binary classification case.  The proposed method reaches a high rank correlation ($0.93$) faster than other two methods but the convergence score is a bit lower than other two methods after more labels have been queried.  However, it doesn't necessarily means our algorithm will perform worse in selecting good labelers since the key is having an accurate estimate of the top labelers.  Our aggressive sampling approach does not do a good job of evaluating those bad labelers whose $\alpha$ value is also an important contributor in the rank correlation score. We will discuss this phenomenon in more depth during the discussion.

\begin{figure}[!htb]
\includegraphics[width=1\columnwidth]{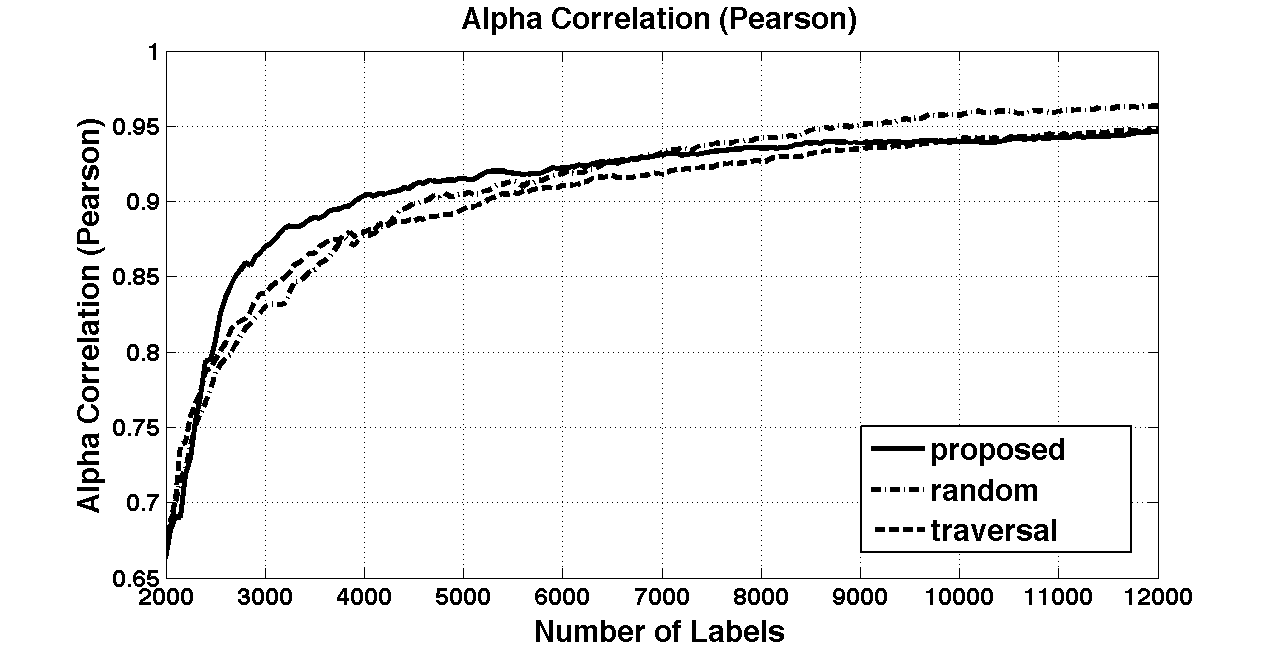}
\caption{Comparison of the different sampling strategies at estimating worker performance in the simulated pool with Gaussian worker performance levels using Pearson correlation.}
\label{fig:exp2pea}
\end{figure}

\begin{figure}[!htb]
\includegraphics[width=1\columnwidth]{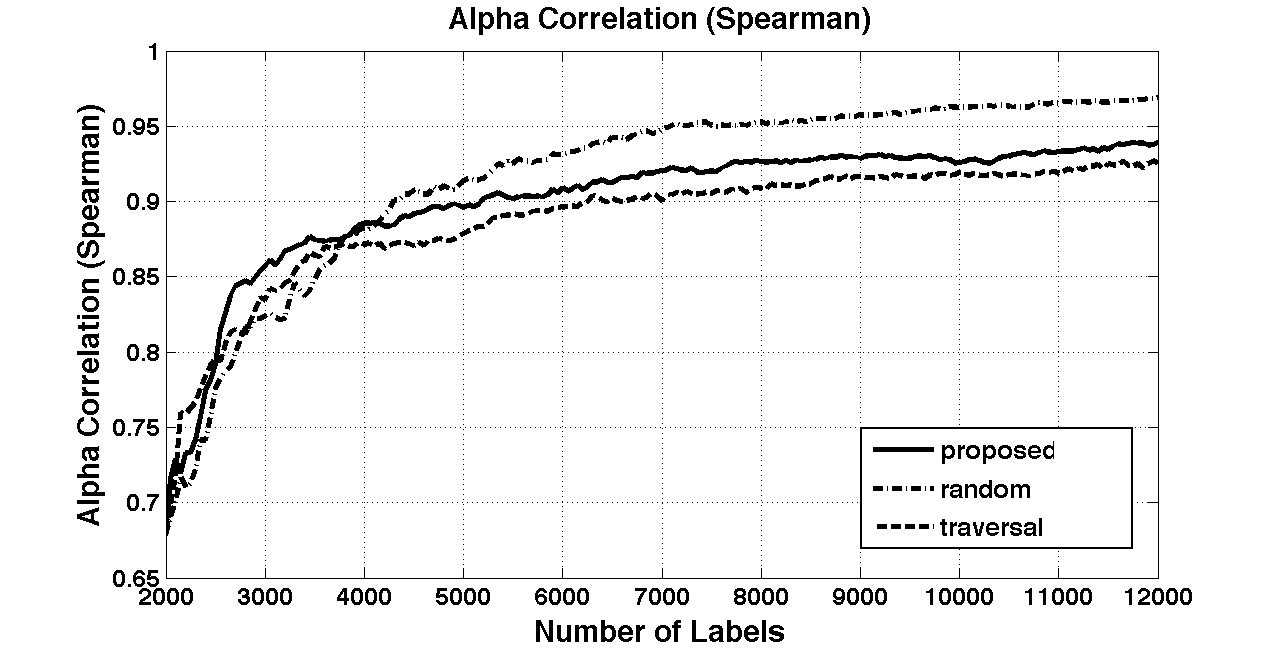}
\caption{Comparison of different sampling strategies at estimated worker performance in the simulated pool with Gaussian worker performance levels using Spearman correlation.}
\label{fig:exp2spe}
\end{figure}

Figure~\ref{fig:exp2peaworker} and Figure~\ref{fig:exp2speworker} show the result of using Pearson and Spearman rank correlation to evaluate the performance of worker selection strategies. As $\alpha'$ and $\beta'$ generated to simulate a more diverse population of labelers and tasks, the weighted selection and $\epsilon$-greedy selection (with $\epsilon = 0.5$) show a faster convergence to a higher correlation score.  The $\epsilon$-greedy selection (with $\epsilon = 0.5$) reaches a high rank correlation ($0.92$) after 2000 iterations and finally converge at $0.96$.

\begin{figure}[!htb]
\includegraphics[width=1\columnwidth]{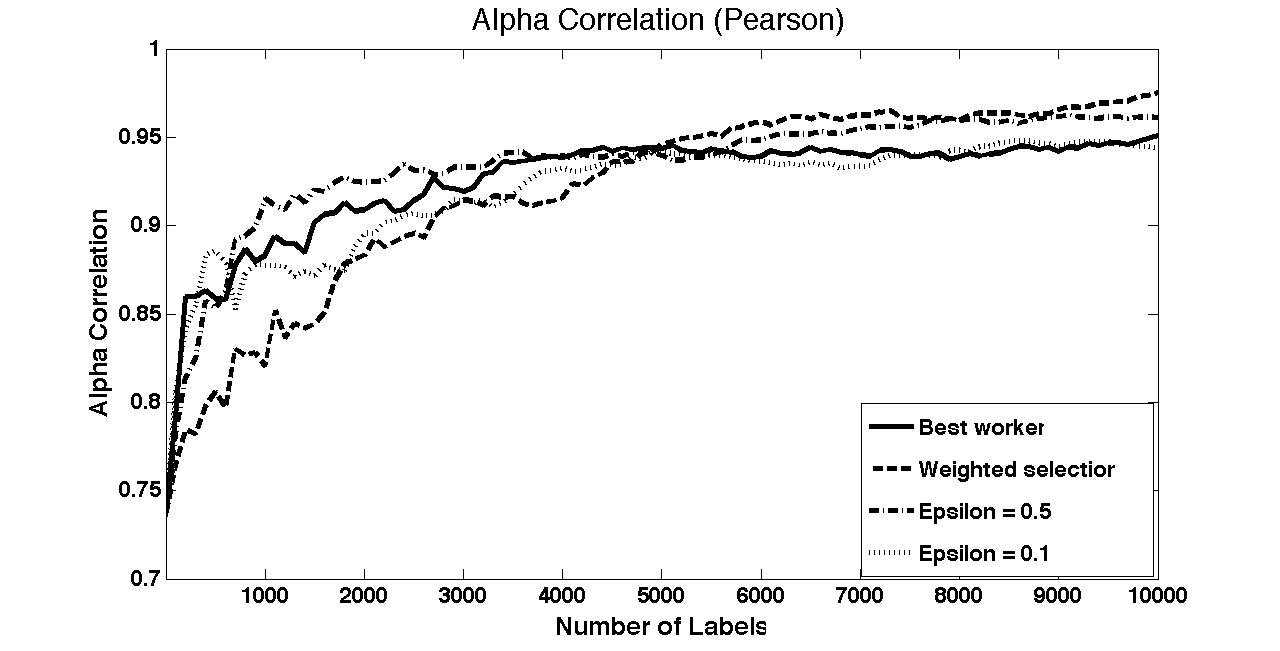}
\caption{Comparison of the different worker selection strategies at estimating worker performance in the simulated pool with Gaussian worker performance levels using Pearson correlation.}
\label{fig:exp2peaworker}
\end{figure}

\begin{figure}[!htb]
\includegraphics[width=1\columnwidth]{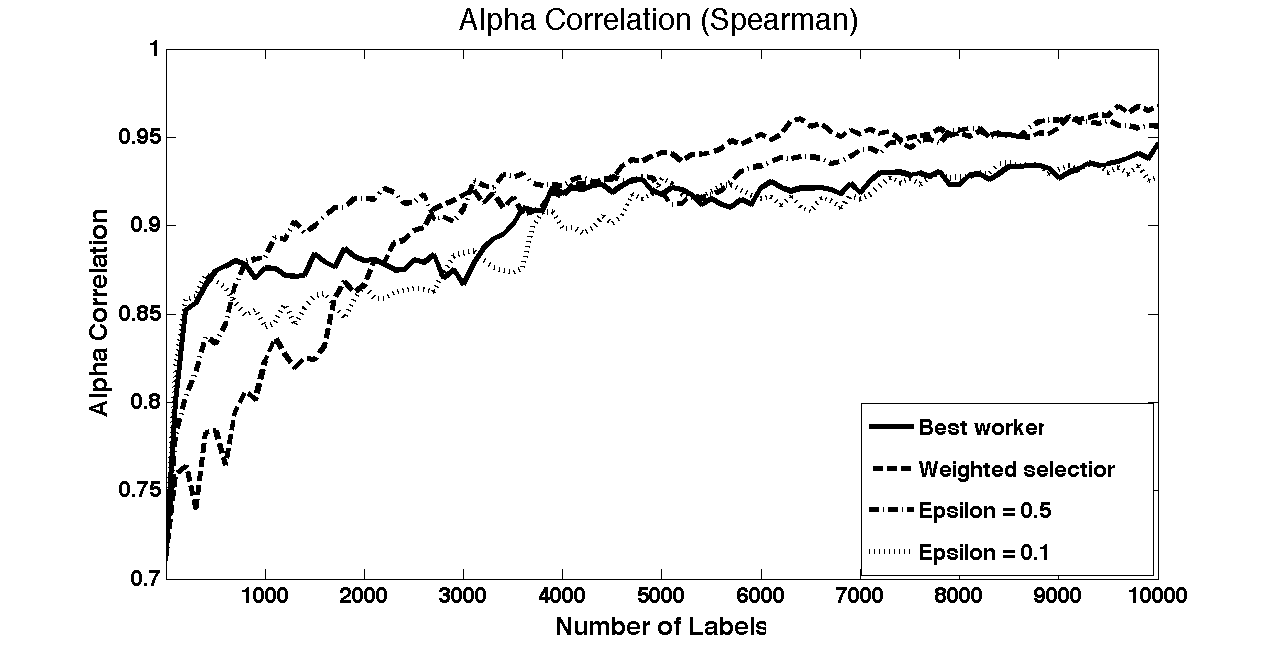}
\caption{Comparison of different worker selection strategies at estimated worker performance in the simulated pool with Gaussian worker performance levels using Spearman correlation.}
\label{fig:exp2speworker}
\end{figure}

\subsubsection{Dataset Crowdsourced with Human Workers}

To evaluate our active learning strategy on a standard binary classification task benchmark, we used a facial image dataset crowdsourced with human labelers on Amazon's Mechanical Turk.  Whitehill et al.~\cite{whitehill2009whose} asked $20$ real human workers on Mechanical Turk to annotate 160 facial images by labeling them as either Duchenne or Non-Duchenne.  A Duchenne smile (enjoyment smile) is distinguished from a Non-Duchenne (social smile) through the activation of the Orbicularis Oculi muscle around the eyes, which the former exhibits and the latter does not.  The dataset consists of $3572$ labels.  The real worker may provide more than one label with opposite results  to the same task. To evaluate the performance of these workers, these images were also annotated by two certified experts in the Facial Action Coding System. According to the expert labels, 58 out of 160 images contained Duchenne smiles.

\begin{figure}[!htb]
\includegraphics[width=1\columnwidth]{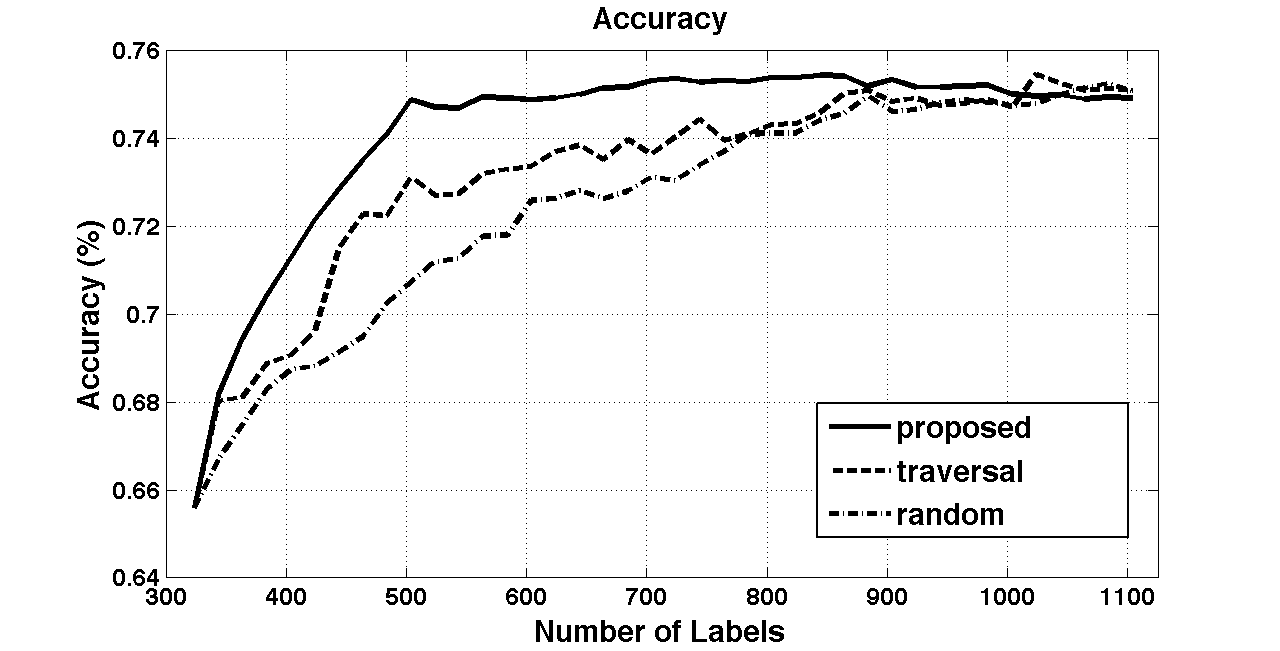}
\caption{Comparison of annotation accuracy with different sampling strategies (proposed, traversal, random) on the facial image dataset crowdsourced on Mechanical Turk.}
\label{fig:exp3acc}
\end{figure}

In this experiment with labels from real MTurk workers, we also initialized the experiment by asking two people to annotate each task, which means there are $160\times2=320 (10\%)$ labels before all strategies start running. Each strategy runs for 800 $(22.7\%)$ iterations.  Figure~\ref{fig:exp3acc} shows the training accuracy with different sampling strategies. Our proposed method rises quickly and converges to $75\%$ training accuracy after soliciting 500 $(14.0\%)$ labels in total. Both traversal and random strategies converge to the same level with 1000 $(28\%)$ labels, which would result in twice the labeling cost of our proposed method.

Figure~\ref{fig:exp3accworker} shows the training accuracy with different worker selection strategies. The experimental results indicate that the performance of $\epsilon$-greedy selection algorithm which rises quickly and converges to $76\%$ training accuracy after soliciting 200 $(5.6\%)$ labels in total is comparable good as the best worker strategy. However, although we run 10 times for each strategy, the performance of weighted selection strategy is very unstable with real crowdsourced annotations.

\begin{figure}[!htb]
\includegraphics[width=1\columnwidth]{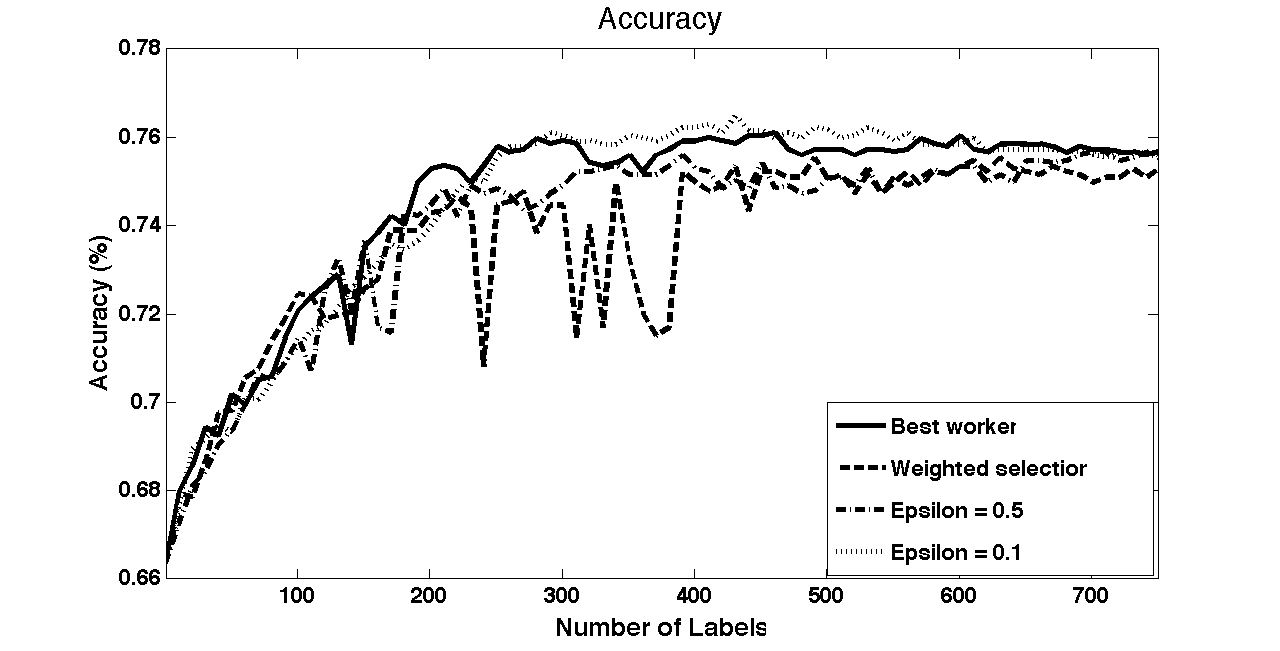}
\caption{Comparison of annotation accuracy with different worker selection strategies on the facial image dataset crowdsourced on Mechanical Turk.}
\label{fig:exp3accworker}
\end{figure}

\section{Discussion}

In real-world crowdsourcing applications, annotation accuracy and budget limitations are obviously the most important immediate criteria for evaluating the performance of a learning model.  However, identifying knowledgeable and reliable workers is potentially useful because these workers could be employed in future annotation tasks. The difficulty of tasks mainly serves as a discriminant for distinguishing between good and bad labelers, rather than a evaluation score for how well the learning model performs. Especially in active learning, aggressive sampling is unable to perform well on all criteria so task difficulty should be sacrificed for performance gains on the other metrics.

Our algorithm aggressively opts to use the best workers possible which yields labeling improvements but makes it difficult to accurately assess the relative rank of the poor labelers.  We believe that a simple analysis of the rank correlation may not the best way to evaluate the estimation $\pmb{\alpha}'$ since in most real-world applications labelers under a certain performance level should be eliminated early.  The pool of potential workers that can be reached using MTurk is very large so devoting annotation budget to working with poor labelers is unecessary.  Our proposed method is good at dividing labelers into two groups (good and bad) which enables it to perform well in the first simulation experiment while failing to make the subtle discriminations between relatively poor labelers required for assessing worker performance in the second experiment.  However, in practical crowdsourcing tasks, filtering out bad labelers is enough for collecting reliable labels in future tasks, and the bottom ranked workers are largely irrelevant to the overall performance of the crowdsourcing pipeline.

\section{Conclusion}

Although crowdsourcing annotations using active learning is an attractive and affordable idea for large-scale data labeling, the approach poses significant difficulties.  Several studies in different research domains show that active learning approaches developed for noise-free annotations do not perform well with crowdsourced data.  This paper presents a practical approach for using active learning in conjunction with Bayesian networks to model both the expertise of unknown labelers and the difficulty of annotation tasks. 

Our work makes two contributions that enable us to robustly train a probabilistic graphical model under these challenging conditions. First, we propose an original and efficient sampling criteria which iteratively assigns the most reliable labelers to the tasks with the highest labeling risk. Second, we present comprehensive evaluations on both simulated and real-world datasets that show the strength of our proposed approach in significantly reducing the quantity of labels required for training the model.  Our experiments using crowdsourced data from Mechanical Turk confirm that the proposed approach improves active learning in noisy real-world conditions.

\newpage

\bibliographystyle{plain}
\bibliography{arxiv}

\end{document}